\documentclass[11pt]{article} 
\usepackage{rldmsubmit,palatino}
\usepackage{multirow}
\usepackage{graphicx}
\usepackage{subcaption}
\usepackage{bm}
\usepackage{amsmath}
\usepackage[style=numeric,sorting=none]{biblatex} 
\title{Modularity benefits reinforcement learning agents with competing homeostatic drives}

\author{
Zack Dulberg  \\
Princeton University\\
Princeton, NJ 08544 \\
\texttt{zdulberg@princeton.edu} \\
\And
Rachit Dubey  \\
Princeton University\\
Princeton, NJ 08544 \\
\texttt{rdubey@princeton.edu} \\
\And
Isabel M. Berwian  \\
Princeton University\\
Princeton, NJ 08544 \\
\texttt{iberwian@princeton.edu} \\
\And
Jonathan D. Cohen  \\
Princeton University\\
Princeton, NJ 08544 \\
\texttt{zdulberg@princeton.edu} \\
}

\begin{document}

\maketitle

\begin{abstract}

The problem of balancing conflicting needs is fundamental to intelligence. Standard reinforcement learning algorithms maximize a scalar reward, which requires combining different objective-specific rewards into a single number. Alternatively, different objectives could also be combined at the level of \emph{action} value, such that specialist modules responsible for different objectives submit different action suggestions to a decision process, each based on rewards that are independent of one another. In this work, we explore the potential benefits of this alternative strategy. We investigate a biologically relevant multi-objective problem, the continual homeostasis of a set of variables, and compare a monolithic deep Q-network to a modular network with a dedicated Q-learner for each variable. We find that the modular agent: a) requires minimal exogenously determined exploration; b) has improved sample efficiency; and c) is more robust to out-of-domain perturbation. 

\end{abstract}


\acknowledgements{This project / publication was made possible through the support of a grant from the John Templeton Foundation.}

\startmain 

\section{Introduction}
\vspace{-0.2cm}

Humans (and other animals) must satisfy a large set of distinct and possibly conflicting objectives. For example, we must find food, water, shelter, socialize, maintain our temperature, reproduce, etc.. Artificial agents that need to function autonomously in natural environments may face similar problems (e.g., balancing the need to accomplish a specified goal with the need to recharge, etc.). Finding a way to balance disparate needs is thus an important challenge for intelligent agents, and can often be a source of psychological conflict in humans. 

In standard reinforcement learning (RL), monolithic agents act in order to maximize a single future discounted reward \cite{sutton2018reinforcement}. The standard way to generalize this to problems with multiple objectives is scalarization. For example, in homeostatically-regulated reinforcement learning (HRRL), an agent is modelled as having separable homeostatic drives, and is rewarded based on its ability to maintain all its "homeostats" at their set points.  This is done by \emph{combining} deviations from \emph{all} set-points into a single reward which, when maximized, minimizes homeostatic deviations overall \cite{keramati2014homeostatic}. 


This approach faces several challenges typical to RL. First, to avoid settling on a sub-optimal policy, an agent must trade-off exploitation of knowledge about its primary objective with some form of exogenous exploration (typically by acting randomly or according to an exploration-specific bonus). Second, sample inefficiency follows from the “curse of dimensionality”: as environmental complexity increases, an agent must learn how exponentially more states relate to its objective. Third, RL agents tend to over-fit their environment, performing poorly out-of-domain (i.e. they are not robust to distribution shifts). In the broader context of balancing multiple objectives, reward scalarization might be undesirable if the relative importance of different objectives is unknown or variable \cite{roijers2013survey}. Finally, is not clear that the brain itself uses a common currency to navigate such trade-offs \cite{hayden2021case}. 

What is the alternative? Given that objectives may conflict with each other due to environmental constraints, and that agents only have one body with which to act, conflict must be resolved at some point between affordance and action. The monolithic solution resolves conflict at the level of reward (i.e. close to affordance). We suggest resolution could occur later; a set of modules with separate reward functions could submit action values to a decision process that selects a final action \cite{russell2003q,van2017hybrid}. This approach has the potential to address the three aforementioned challenges. Exploration might emerge naturally as a property of the system rather than having to be imposed or regulated as a separate factor, as specialist modules are ``dragged along" by other modules when those have the ``upper hand" on action. Modules might also have smaller sub-sets of relevant features to learn about, improving sample efficiency, and be less sensitive to distribution shifts in irrelevant features, improving robustness.

Here, we report simulations that provide evidence for benefits of such a modular approach with respect to exploration, sample efficiency, and robustness using deep RL in the context of homeostatic objectives. We construct a simple but flexible environment of homeostatic tasks and construct a deep RL implementation of the HRRL reward function. We then use this framework to quantify differences between monolithic and modular deep Q-agents, finding that modular agents seem to explore well on their own, achieve homeostasis faster, and better maintain it after distribution shift.  Together, these results highlight the potential learning benefits of modular RL, while at the same time offering a framework through which psychological conflict and resolution might be better understood. 



\vspace{-0.2cm}

\section{Methods}
\vspace{-0.2cm}
\subsection{Environment}

\begin{figure}[t!]
\centering
\begin{subfigure}[t]{0.45\linewidth}
\includegraphics[width=\linewidth]{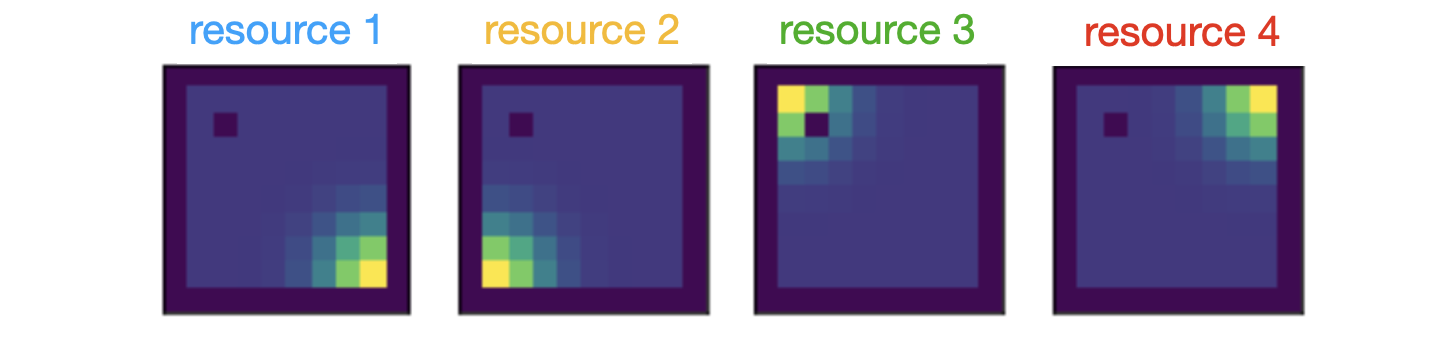}  
\subcaption{Gridworld environment}
\label{res}
\end{subfigure}
\begin{subfigure}[t]{0.5\linewidth}
\includegraphics[width=\linewidth]{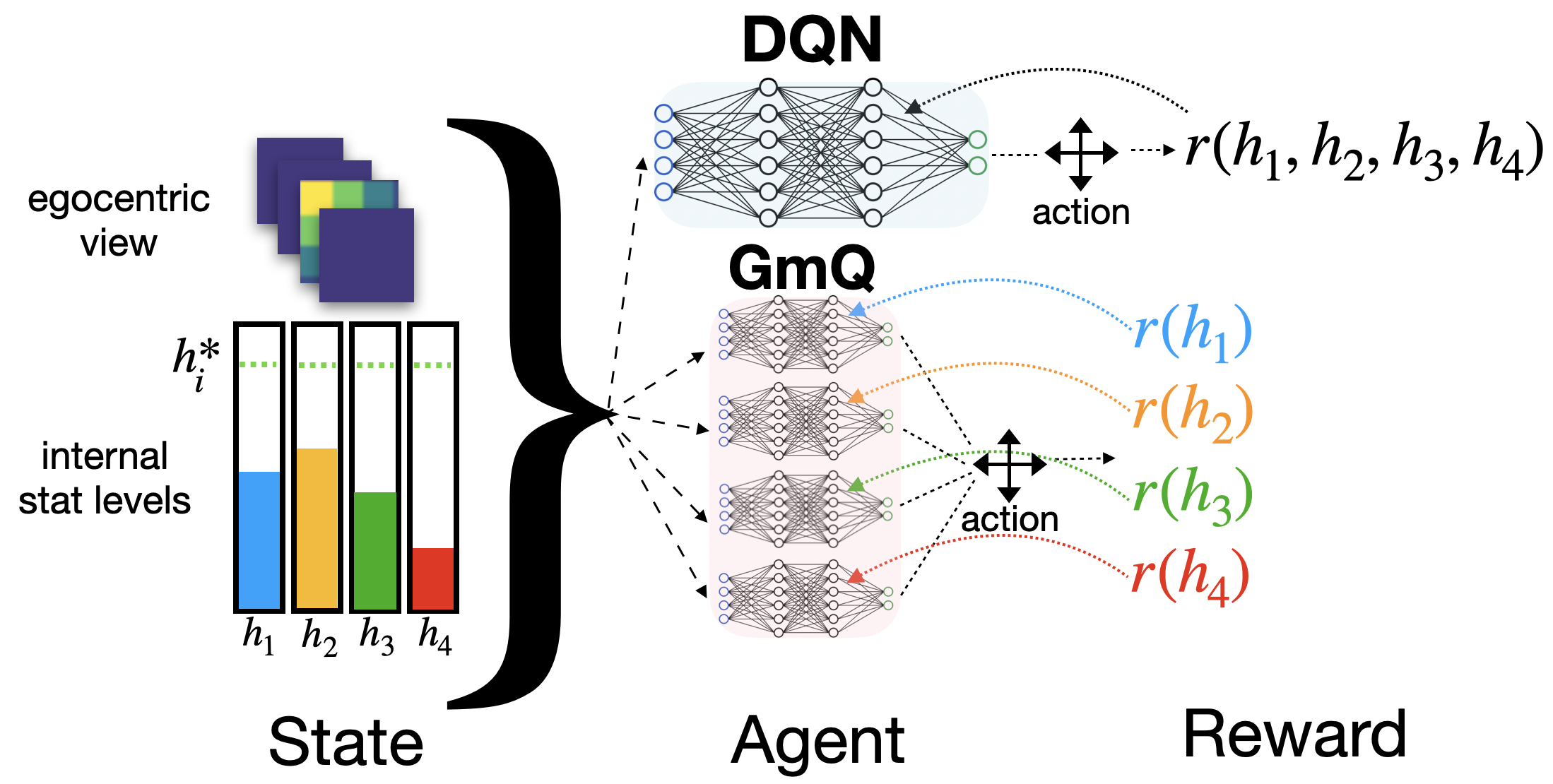} 
\subcaption{Overview of monolithic DQN and modular GmQ agents}
\label{scheme}
\end{subfigure}
\caption{Environment and model schematics}
\vspace{-0.4cm}
\label{resource}
\end{figure}

To study conflicting needs, we constructed a toy grid-world environment containing multiple different resources. Specifically, each location $(x,y)$ in the environment contained a vector of resources of length $N$ (i.e., there were $N$ overlaid resource maps). The spatial distribution of each individual resource was specified by a normalized 2D Gaussian with mean $\mu_x, \mu_y$ and co-variance matrix $\Sigma$ (see also Figure \ref{res}).

The agent received as perceptual input a 3x3 egocentric slice of the $N$ resource maps (i.e. it could see all the resource levels at each position in its local vicinity). In addition to the resource landscape, the agent also perceived a vector $H_t = (h_{1,t}, h_{2,t},...,h_{N,t})$ consisting of $N$ internal variables with each representing the agent's homeostatic need with respect to the resource. We refer to these variables as "internal stats" or just "stats" (such as osmostat, glucostat, etc.) which we assume are independent ($h_i$ is only affected by acquisition of resource $i$) and have some desired set-point $h_i^*$ (see Figure \ref{scheme}). Set-points were fixed at $H^* = (h_{1}^*, h_{2}^*,...,h_{N}^*)$ and did not change over the course of training.

The agent could move in each of four cardinal directions and, with each step, the individual stats, $h_i$, increased by the amount of resource $i$ at the agent's next location. Additionally, each internal stat decayed at a constant rate to represent the natural depletion of internal resources over time (note: resources in the environment themselves did not deplete). Thus, if the agent discovered a location with a high level of resource for a single depleted stat, staying at that location would optimize that stat toward its set-point, however others would progressively deplete.  Agents were initialized in the center of the grid, with internal stats below their set-points, and were trained for $30,000$ steps for a single episode (i.e. agents had to learn in real time as internal stats depleted).  For all experiments, the internal stats started at the same level and shared the same set-points. Environmental parameters are summarized in Table \ref{hyper}.

\subsection{Models}

\paragraph{Monolithic agent} We created a monolithic agent based on the deep Q network (DQN) \cite{mnih2013playing}. The agent's perceptual input was a concatenation of all local resource levels along with all internal stat levels at each time step (we used neural networks as function approximators since stats were continuous variables). It's output was 4 action logits subsequently used for $\epsilon$-greedy action selection. We used the HRRL reward function \cite{keramati2014homeostatic} which defined reward at each time-step $r_t$ as drive reduction, where drive $D$ was a convex function of set-point deviations; see equation (\ref{reward}). 

\begin{equation}
\vspace{-0.2cm}
r_t = D(H_t) - D(H_{t+1})
\quad\text{where}\quad
D(H_t) = \sqrt[m]{\sum_{i=1}^N | h_i^* - h_{i,t} |^n }
\label{reward}
\end{equation}


\paragraph{Modular agent} We created a modular agent based on greatest-mass Q-learning (GmQ) \cite{russell2003q}, which consisted of a separate DQN for each of the 4 resources/stats. Here, each module had the same input as the monolithic model (i.e. the full egocentric view and all 4 stat levels), but received a separate reward $r_{i,t}$ derived from only a single stat. The reward function for the $i$th module was therefore defined as in equation (\ref{1Dr}), where drive $D$ depended on the $i$th resource only. 

\begin{equation}
\vspace{-0.2cm}
r_{i,t} = D(h_{i,t}) - D(h_{i,t+1})
\quad\text{where}\quad
D(h_{i,t}) = \sqrt[m]{|h_i^* - h_{i,t} |^n }
\label{1Dr}
\end{equation}


To select a single action from the suggestions of the multiple modules, we used a simple additive heuristic.  We first summed Q-values for each action across modules, and then performed standard $\epsilon$-greedy action selection on the result. More specifically, if $Q_i(a)$ was the Q-value of action $a$ suggested by module $i$, greedy actions were selected as $\underset{a}{\arg\max} \sum\limits_i Q_{i}(a)$.

\paragraph{Common features} Schematics for both models are shown in Figure \ref{scheme}. All Q-networks were multi-layered perceptrons (MLP) with rectified linear nonlinearities trained using a standard temporal difference loss function with experience replay and target networks \cite{mnih2013playing}. The Adam optimizer was used to perform one gradient update on each step in the environment. For both models, $\epsilon$ was annealed linearly from its initial to final value at the beginning of training at a rate that was experimentally manipulated as described below. Hyperparameters are summarized in Table \ref{hyper}.

\begin{table}[ht]
\caption{Parameter settings for environment and models}
\vspace{-0.4cm}
\centering
\begin{tabular}{l|ll|l|l}\\
 Model & DQN & GmQ & Environment & \\
\hline
\hline 
Trainable parameters & 1.09e6 & 1.09e6 & \# of resources $N$ & 4\\
MLP hidden layer units & 1024 & 500 & HRRL exponents $(n, m)$ & $(4, 2)$\\
Learning rate & 1e-3 & 1e-3 & Stat set-points $H^*$ & $(5,5,5,5)$ \\
Discount factor $\gamma$ & 0.5 & 0.5 & Initial stat levels $H_{t=0}$ & $(0.5,0.5,0.5,0.5)$ \\
Memory buffer capacity & 30k & 30k & Resource locations $\mu_x, \mu_y$ & \{0,10\},\{0,10\}\\
Target network update frequency & 200 & 200 & Resource covariance $\Sigma$ & 

$\big(\begin{smallmatrix}
  1 & 0\\
  0 & 1
\end{smallmatrix}\big)$
\\
Batch size & 512 & 512 & Stat depletion per step & 0.004
\\
Initial $\epsilon$ & 1 & 1  \\
Final $\epsilon$ & 0.01 & 0.01 \\
\end{tabular}
\vspace{-0.35cm}
\label{hyper}
\end{table}

\section{Results}
\vspace{-0.3cm}
\subsection{Optimizing monolithic DQN for homeostasis}

\begin{figure}[b!]
\vspace{-0.3cm}
\centering
\begin{subfigure}[t]{0.45\linewidth}
    \centering
    \includegraphics[width=0.55\linewidth]{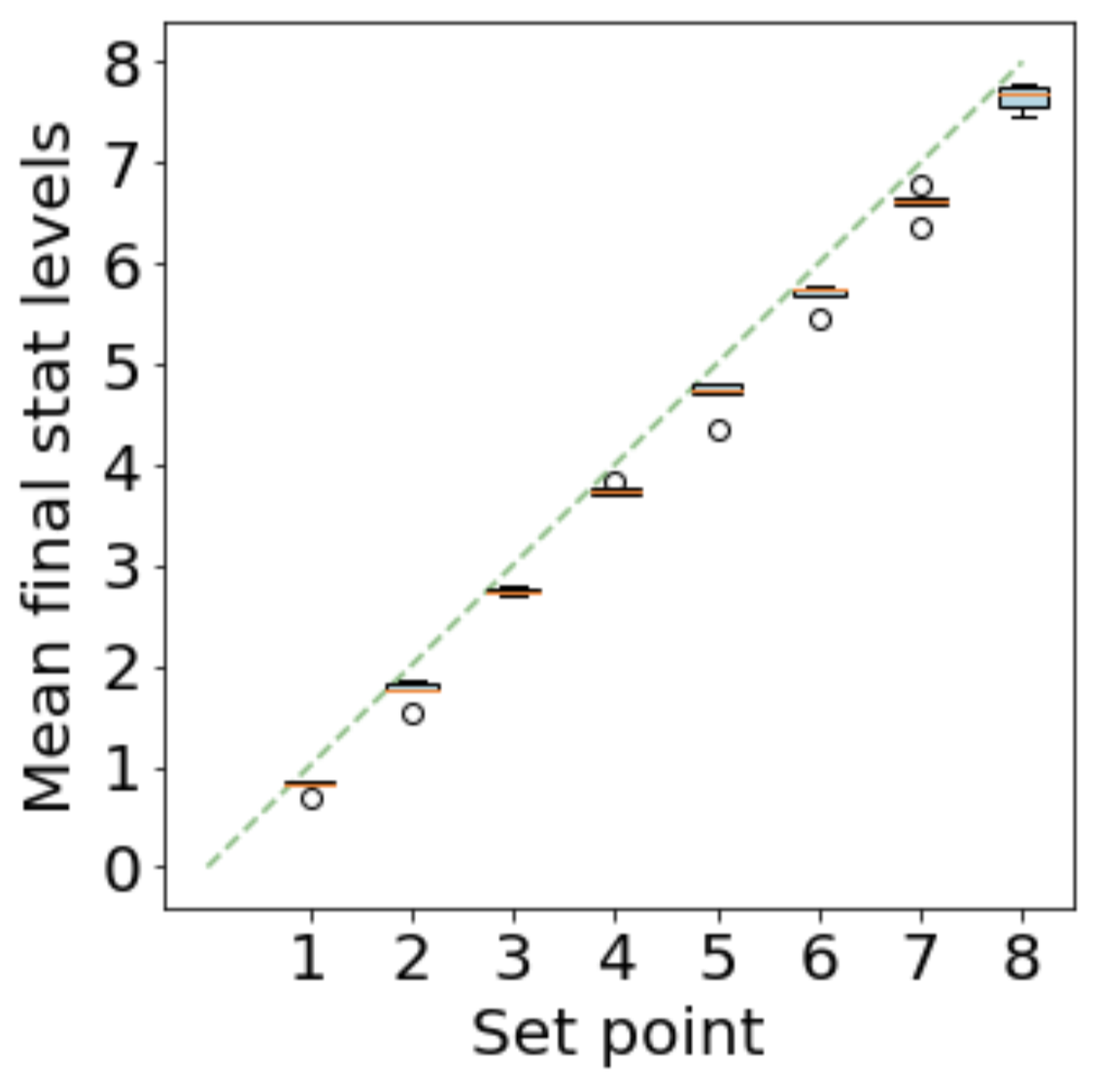}  
    \caption{Homeostasis is achieved for a range of set-points ($n=10$)\vspace{-0.2cm}}
    \label{setpoint}
\end{subfigure}\hfill
\begin{subfigure}[t]{0.45\linewidth}
    \centering
    \includegraphics[width=\linewidth]{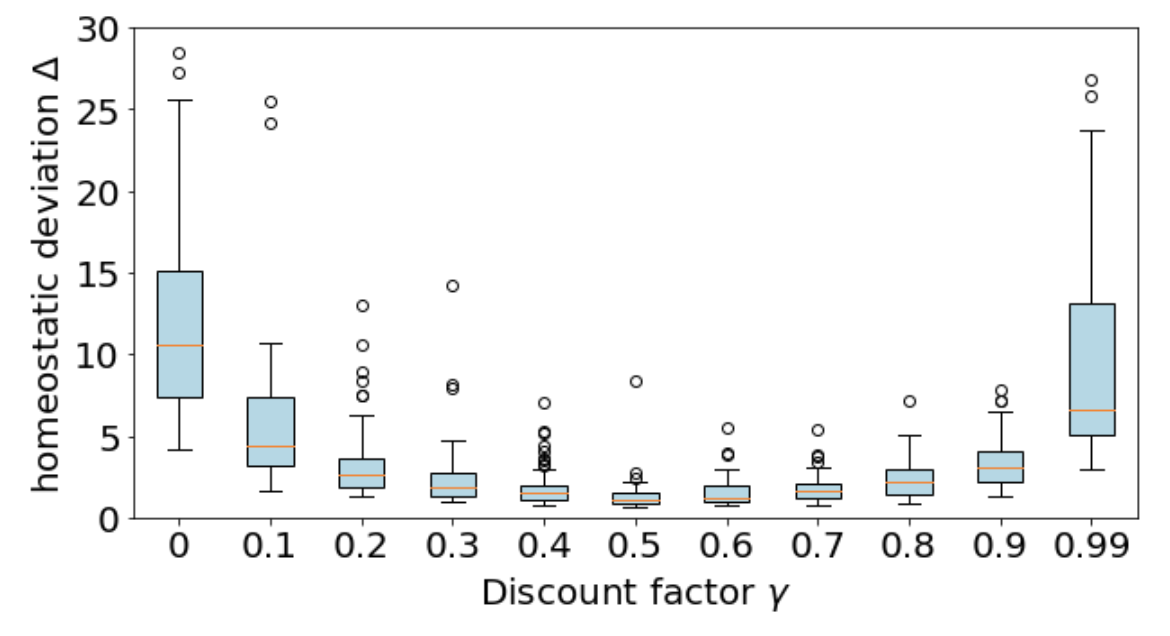}
    \caption{Performance over a range of discount factors ( $n=50$)\vspace{-0.2cm}}
    \label{gamma}
\end{subfigure}
\caption{DQN baseline learnability and performance; Boxplots display inter-quartile range and outliers for $n$ models}
\label{dqn}
\end{figure}

We first characterized whether a standard DQN could reliably perform the task of homeostasis in our environment. Figure \ref{setpoint} summarizes the mean internal stat levels of 10 models averaged over all 4 stats and over the final 1k steps of training for different desired set-points. The model reliably achieved each set-point by the end of training, indicated by points tracking the identity line. The slight under-shooting (i.e. points slightly below the diagonal line) may reflect a bias from stats being initialized far below their set-points. 

We then fixed the set-points $h_i^* = 5$ for all stats, and optimized the performance of the DQN baseline by performing a search over performance-relevant hyper-parameters, such as the discount factor $\gamma$. To quantify performance, we calculated the average homeostatic deviation per step $\Delta$ after the exploration annealing phase (using $t_1 = 15k$ and $t_2 = 30k$) as in equation (\ref{delta}). Lower $\Delta$ indicates better homeostatic performance.\vspace{-0.2cm}

\begin{equation}
\vspace{-0.2cm}
\Delta = \frac{\sum\limits_{t=t_1}^{t_2} \sum\limits_i | h_i^* - h_{i,t} |}{t_2 - t_1}
\label{delta}
\end{equation}

Baseline DQN performance over a range of discount factors $\gamma$ is shown in Figure \ref{gamma}. We selected the best performing setting of $\gamma = 0.5$ and matched this and other parameters (see Table \ref{hyper}) between DQN and GmQ to compare them in the following two head-to-head experiments. 

\subsection{Modularity provides an exploration benefit}

To investigate the impact of modularity on the need for exploration, we systematically varied the number of steps used to anneal $\epsilon$ in $\epsilon$-greedy exploration from its initial to final value. We varied the $\epsilon$ annealing time for both models from 1 (i.e. minimal exploration of $\epsilon = 0.01$ only) to 10k (i.e. annealing from $\epsilon = 1$  to $\epsilon = 0.01$ over 10k steps). 

Figure \ref{explore} shows the results of varying the amount of exploration annealing for both models. While DQN gains incremental performance benefits from increasing periods of initial exploration, GmQ displays a striking indifference to the exploration period; with only 1 step of annealing, it performs as well or better than the best DQN models. In other words, DQN requires careful tuning of an appropriate exploration annealing period, but GmQ achieves good performance with effectively no exogenously specified exploration.


\subsection{Modularity provides robustness in the face of perturbation}

Finally, we tested how robust each model was to a perturbation out-of-domain that occurred halfway through training, i.e. at time-step 15k. At that point, a single internal stat variable (i.e. $h_4$) was clamped to a value of 20 (a value previously unseen by the network), and did not change (thus contributing no drive reduction and therefore 0 reward). We tested how well homeostasis was maintained for remaining stats after this perturbation. 

\begin{figure}[t!]
\centering
\begin{subfigure}[t]{0.5\linewidth}
  \centering
  \includegraphics[width=0.77\linewidth]{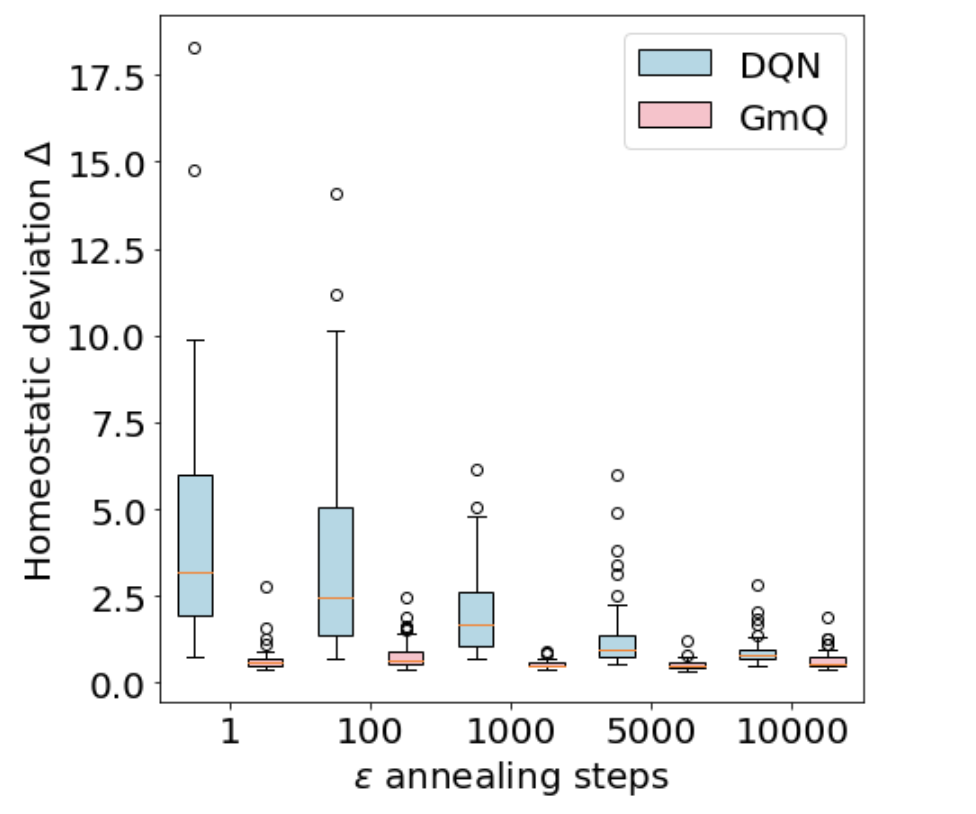}  
  \vspace{-0.1cm}
  \subcaption{Varying exploration period ($n=50$)\vspace{-0.2cm}}
  \label{explore}
\end{subfigure}%
\begin{subfigure}[t]{0.5\linewidth}
  \centering
  \includegraphics[width=0.62\linewidth]{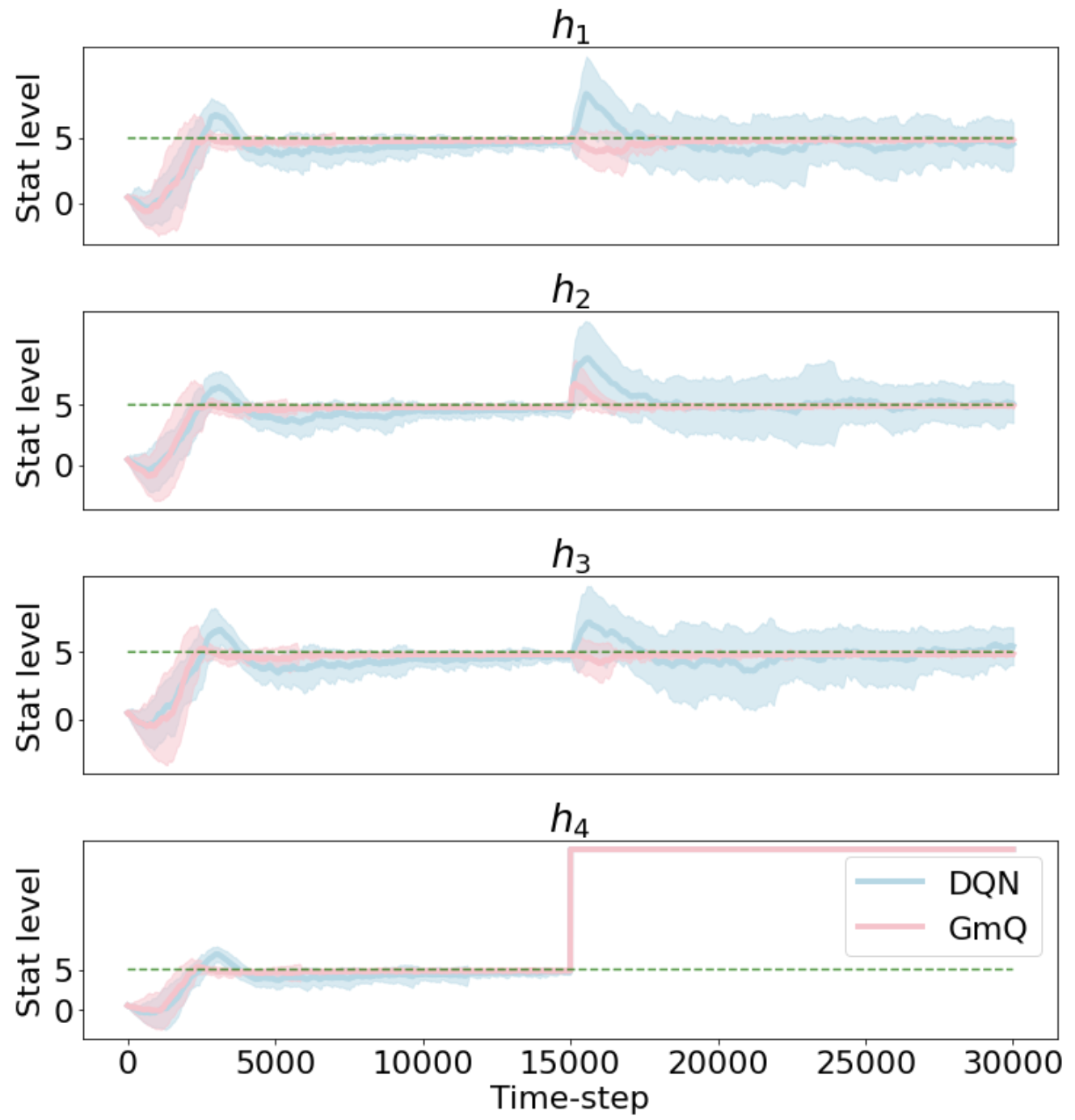}
  \vspace{-0.1cm}
  \subcaption{Stat time-courses ($n=50$) perturbed at $t=15k$ by clamping $h_4$. Green line shows set-point. Shading reflects s.d. accross models.\vspace{-0.2cm}}
  \label{perturb}
\end{subfigure}
\caption{Experiments comparing DQN and GmQ with respect to (a) exploration and (b) perturbation}
\vspace{-0.3cm}
\label{experiments}
\end{figure}

Figure \ref{perturb} shows the time-course of the four stats from the beginning of training, through a perturbation at time-step 15k, using $5000$ $\epsilon$-annealing steps. First, it can be seen that GmQ achieves stable homeostasis first, whereas DQN over-shoots set-points initially and takes longer to stabilize. Second, when stat 4 is clamped, DQN displays a significant disturbance to homeostasis of remaining stats, without clear recovery, whereas GmQ is robust in the face of the perturbation.

\section{Discussion}
\vspace{-0.2cm}
We have shown that in a grid-world task with competing homeostatic drives, a simple modular agent based on greatest-mass Q-learning (GmQ) requires less hand-coded exploration, learns faster, and is more robust to environmental perturbations compared to a traditional monolithic deep Q-network (DQN). Our findings in the context of competing drives complement work showing mixture of expert systems display improved sample efficiency and generalization \cite{jacobs1991adaptive}. We also believe the exploration benefits we observed are novel. The problem of exploration in RL is fundamental, and existing solutions make use of noise, explicit exploratory drives/bonuses, and/or other forms of auto-annealing \cite{yang2021exploration,mcclure2005exploration}. We suggest an additional class of strategies, namely, exploration as an added benefit of having multiple independent drives, since exploitation from the perspective of one module is exploration from the perspective of another. We hypothesize that the ability of modules to suggest conflicting actions may provide modular agents with an implicit source of exploration. 
\vspace{-0.3cm}
\paragraph{Future Work} Our toy environment has highlighted some initial benefits of modularity (exploration, sample efficiency and robustness), and we predict that these advantages will be amplified in more complex environments, or as the number of drives/objectives increases, due to the curse of dimensionality (more states to explore, learn about, or perturb). We aim to test our agents in rich 3D environments with homeostatic objectives. Next, modular drives immediately pose the problem of coordination. While we simply summed Q-values, more complex arbitrators (such as an additional RL agent that dynamically re-weights individual drives) might better exploit the benefits of modularity in the context of multiple objectives. Finally, humans experience psychological conflict, with various resolution mechanisms long described by psychodynamic theories  \cite{freud2018ego}. Modular RL, with its implicit conflicts and resolutions, could, for the first time, offer a formal, computationally-explicit, and normative explanatory framework that could undergird and/or replace elements of psychodynamic theory for understanding the mechanisms responsible for conflict and resolution in the human brain.

\printbibliography 

\end{document}